\documentclass[letterpaper, 10 pt, conference]{ieeeconf}  

\IEEEoverridecommandlockouts                              

\usepackage{cite}
\usepackage{graphics} 
\usepackage{epsfig} 
\usepackage{mathptmx} 
\usepackage{amsmath} 
\usepackage{amssymb}  
\usepackage{acronym}
\usepackage{hyperref}

\newacro{CNN}{Convolutional Neural Network}
\newacro{RNN}{Recurrent Neural Network}
\newacro{RL}{Reinforcement Learning}
\newacro{PPO}{Proximal Policy Optimization}
\newacro{WFC}{Wave Function Collapse}
\newacro{CPU}{Central Processing Unit}
\newacro{GPU}{Graphics Processing Unit}

\title{\LARGE \bf
Learning to walk in confined spaces using 3D representation
}

\author{Takahiro Miki, Joonho Lee, Lorenz Wellhausen and Marco Hutter  
\thanks{This project has received funding from the European Union’s Horizon 2020 research and innovation programme under grant agreement No 780883.}
\thanks{This research was supported by the Swiss National Science Foundation (SNSF) as part of project No.188596.}
\thanks{This project has received funding from the European Union’s Horizon 2020 research and innovation programme under grant agreement No 101016970. 
}
\thanks{All authors are with Robotic Systems Lab, ETH Zurich}%
}

\begin{document}

\maketitle
\begin{abstract}
Legged robots have the potential to traverse complex terrain and access confined spaces beyond the reach of traditional platforms thanks to their ability to carefully select footholds and flexibly adapt their body posture while walking. However, robust deployment in real-world applications is still an open challenge. In this paper, we present a method for legged locomotion control using reinforcement learning and 3D volumetric representations to enable robust and versatile locomotion in confined and unstructured environments. By employing a two-layer hierarchical policy structure, we exploit the capabilities of a highly robust low-level policy to follow 6D commands and a high-level policy to enable three-dimensional spatial awareness for navigating under overhanging obstacles. 
Our study includes the development of a procedural terrain generator to create diverse training environments. We present a series of experimental evaluations in both simulation and real-world settings, demonstrating the effectiveness of our approach in controlling a quadruped robot in confined, rough terrain. By achieving this, our work extends the applicability of legged robots to a broader range of scenarios.
\end{abstract}

\section{INTRODUCTION}

Robots have attracted attention for their ability to perform a wide range of tasks, including exploring and accessing areas that are dangerous or inaccessible to humans. Among the different robotic platforms, legged robots are well suited for this type of task due to their ability to move on uneven, complex terrain, and to remain stable on unstable surfaces.
In recent years, legged robots have been successfully deployed in a variety of challenging environments, from mountains, forests or simulated space environments~\cite{lee2020learning, miki2022wild, arm2023scientific} to underground spaces such as tunnels and caves~\cite{Tranzatto2022-sm, Tranzatto2022-gf, Bouman2020-gx, Kottege2023-oo}. However, their full potential has yet to be realized. One unique feature of legged robots is their ability to adjust their posture using its high degree of freedom, such as crouching. This provides capabilities to access confined area not easily replicated by other robotic platforms with a similar size. Despite this, the challenge of robustly deploying legged robots in confined or restricted spaces is still an open problem.

Some studies have shown that legged robots can navigate under low obstacles using a special crouching gait. However, these tests were conducted in simple environments such as on flat ground with a simple overhanging obstacle~\cite{Han2023-ze, zhuang2023robot, hoeller2023anymal, Kim2023-nj}.
Buchanan et al. addressed more complex confined spaces including both steps and sloped overhanging obstacles using trajectory optimization techniques for motion planning and a two-layer elevation map to perceive the environment, representing both the ground and the ceiling~\cite{buchanan2021perceptive, Buchanan2019}.
However, the gait was limited to a static gait and it assumed that the map is accurate and the robot would always have clear space to move in the map, which can be a problem if the sensors make errors or provide unclear data.

\begin{figure}[t]
\centering
\includegraphics[width=\linewidth]{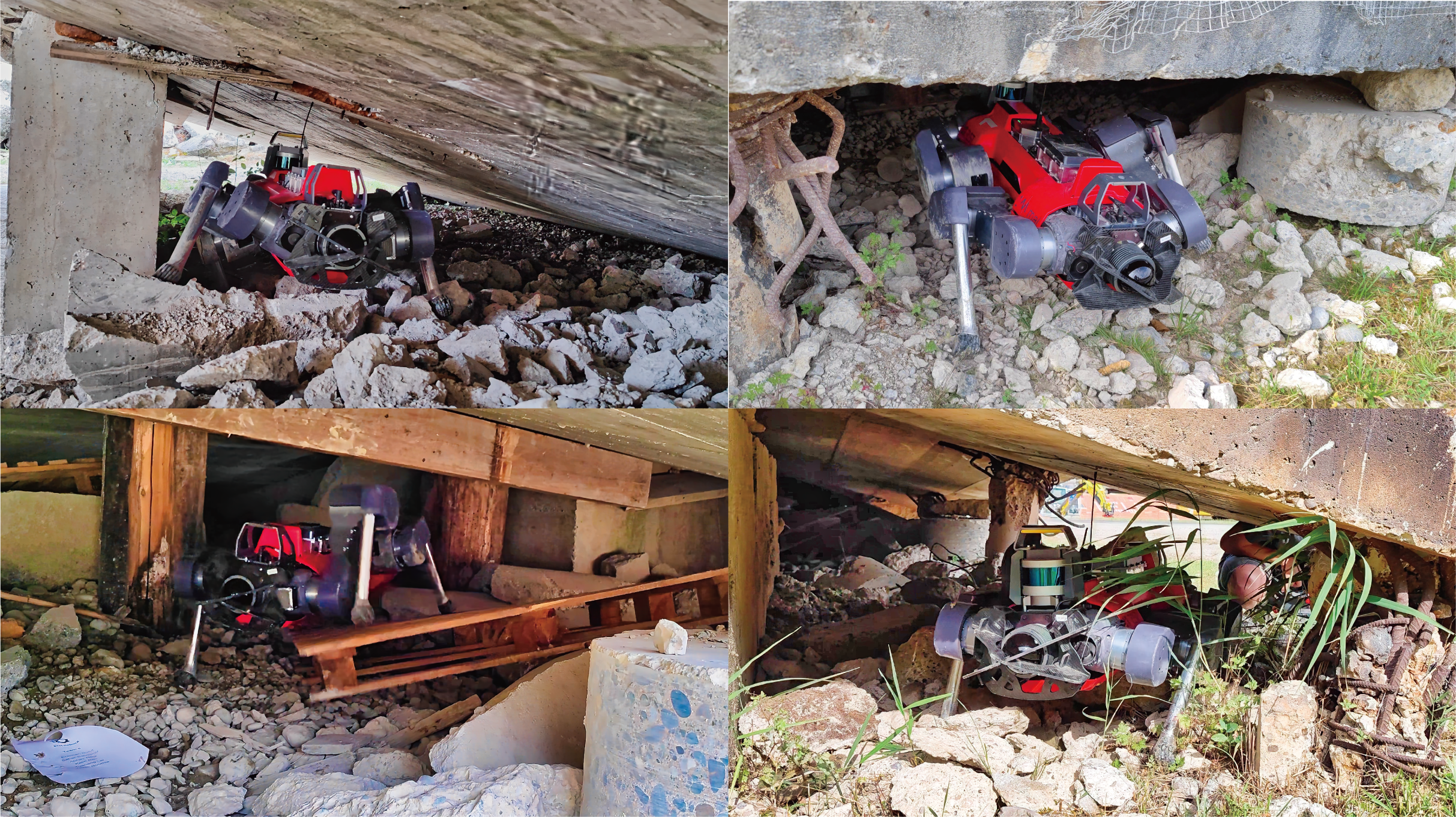}
\caption{Real-world experiment: Successful confined space traversal by the quadruped robot including a simulated collapsed building environment. The terrain consists of loose gravel or unstable steps, while the overhead structures have tilted configurations with narrow openings. The robot could adapt its posture to traverse these challenging conditions.}
\label{fig:deployment}
\end{figure}

To achieve a fast dynamic walking on various terrains while being robust against degraded perception, learning-based methods have recently demonstrated their effectiveness by integrating proprioceptive data with noisy exteroceptive data~\cite{miki2022wild}. However, until now, the deployments are mostly limited to environments without overhanging obstacles~\cite{miki2022wild,yang2022learning,Yu2022-ps,Agarwal2022-hp,Yang2023-tm} or on a flat terrains with single obstacle~\cite{zhuang2023robot, hoeller2023anymal}. 
Our research aims to extend the operational range of legged robots to more complex environments in the wild including confined spaces while keeping the robust and smooth locomotion capability of the state of the art methods.

In this paper, we propose a two-layer hierarchical framework to address the problem. At the lower level, the policy negotiates ground-level obstacles while complying with extended high-level commands such as desired velocity, body height, and body orientation, while the higher level policy uses 3D geometric data to guide the actions of the lower level policy. This architecture allows us to benefit from the robustness of established locomotion methods, while introducing a new layer of spatial awareness through high-level policy formulation.
To achieve spatial awareness of the confined environment, on-board sensors such as cameras, LiDARs, or depth cameras can be used to measure the surrounding geometry. However, they often exhibit degraded performance in the complex environments due to factors such as physical obstructions, blind spots, and limited sensor ranges. In addition, the variety of sensor configurations makes it difficult to create a generalizable policy. To circumvent these challenges, we use a teacher-student training setup where we distill the teacher policy into a robust student policy which takes noisy occupancy voxels as our geometric representation. The voxel representation can integrate different sensor measurements into one data format, allowing a flexibility in different sensor configurations. In addition, the same policy can be used without retraining for the different sensor setup unlike the case of depth images representation. Furthermore, it naturally allows the incorporation of memory-based strategies, such as mapping to preserve information.

The contributions are as follows: First, we introduce a hierarchical policy framework that decomposes the confined space navigation into low-level and high-level tasks where the low-level policy inherits the performance of the robust perceptive locomotion on rough terrain while adding a new capability of entering confined space through high-level policy's posture adaptation. Second, we train the policy in a wide variety of environments by incorporating procedural terrain generator in simulation. Finally, we demonstrate the real-world applicability of our approach through deployments in complex scenarios in the field.

\section{RELATED WORKS}
Environmental perception is essential for legged robots to smoothly navigate complex terrains.
Early methods relied on pre-generated elevation maps for offline planning~\cite{zucker2010optimization, neuhaus2011comprehensive}. This approach evolved to use on-board sensors to dynamically generate elevation maps in real time~\cite{fankhauser2018probabilistic, mikielevation2022, erni2023mem}, which were subsequently used in optimization-based controllers~\cite{kolter2009stereo, havoutis2013onboard, kim2020vision, jenelten2020perceptive, jenelten2022-tq, Grandia2023-mc}. A basic assumption of these methods is accurate terrain information. Therefore, their robustness degrades in the face of mapping inaccuracies, a problem often encountered in unstructured outdoor environments.

On the other hand, learning-based methods have gained momentum for locomotion tasks. 
Reinforcement learning is used to train a neural network policy in simulation and deployed on the hardware~\cite{tan2018sim, Hwangbo2019-jk, lee2020learning, peng2020learning, miki2022wild, rudin2022learning, siekmann2021blind}.
Recently exteroceptive information has been integrated into the locomotion control through elevation map as an intermediate representation to keep past information~\cite{miki2022wild, gangapurwala2020rloc, caluwaerts2023barkour}. They sampled the height scan from the map to feed into a Recurrent Neural Network (RNN) based policy, however this 2.5D representation can not handle overhanging structures.
As another intermediate representation, voxels were used for locomotion task~\cite{Hoeller2022-fw, hoeller2023anymal}. This can represent full 3D structures, however it was demonstrated only in a structured terrain with simple geometry.
Other works have used depth images as input~\cite{imai2022,yang2022learning,Yu2022-ps,Agarwal2022-hp}. In these approaches, depth images are first processed by a \ac{CNN} and either combined with historical features or fed into a \ac{RNN} to preserve past information. 
To extend the memory function of \ac{RNN}-based methods using depth images, implicit voxel representations have been introduced~\cite{Yang2023-tm}. In this method, latent voxels from the previous frame are transformed to the current frame using a predicted transformation obtained from two depth images. However, the verification of these locomotion capabilities did not include diverse and confined spaces.

\section{METHODS}
\subsection{Overview}
\begin{figure*}[htbp]
\centering
\includegraphics[width=0.8\textwidth]{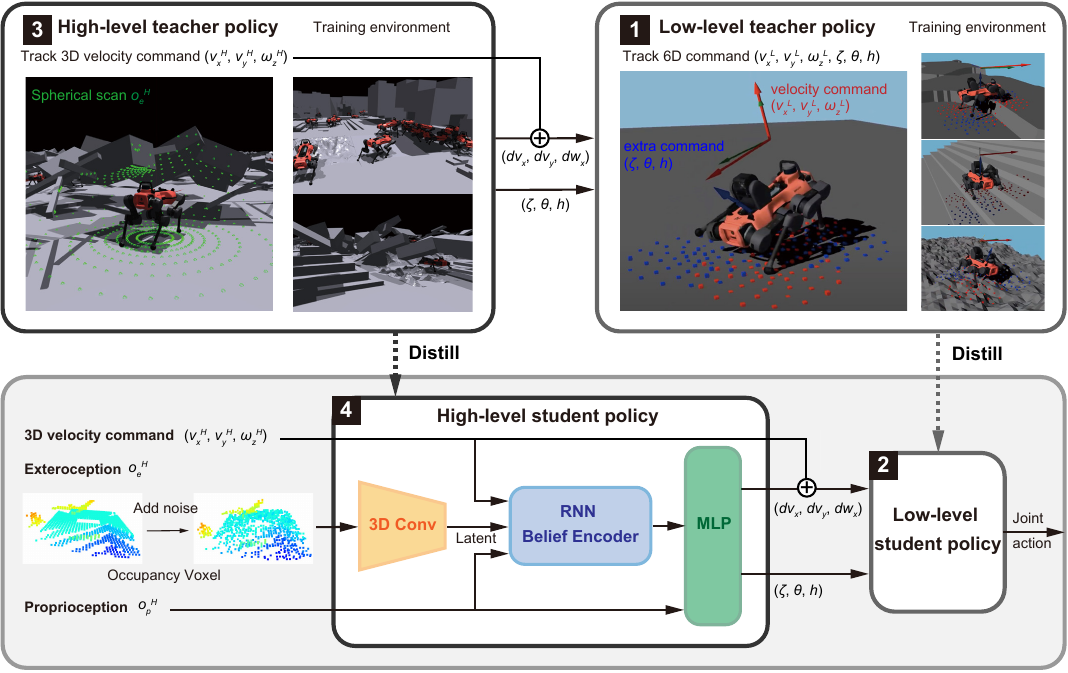}
\caption{Overview of our method. We use a two-layer policy setup.  The low-level policy learns to walk over rough terrain while following 6D commands consisting of x,y, yaw rate and roll pitch and body height.  The high-level policy is trained in a procedurally generated confined environment to guide the robot by giving commands to the low-level policy. We first train a low-level teacher policy and then distill it into a low-level student policy. We follow this by training a high-level teacher policy using spherical scans for exteroceptive perception. Finally, we distill this into a high-level student policy using noisy voxel grids as exteroceptive input.}
\label{fig:overview}
\end{figure*}

Our approach uses a hierarchical framework as shown in Fig. \ref{fig:overview}. The training of the control policy is structured into four progressive stages: the low-level teacher policy, the low-level student policy, the high-level teacher policy, and the high-level student policy.

First, the low-level teacher policy is trained using reinforcement learning to follow high-level commands over varied, rough terrain. Using 6D inputs - including $x$, $y$, and yaw velocities, roll, pitch, and body height - the policy acquires the ability to navigate smoothly over uneven surfaces while following given commands.
Then the low-level student policy is distilled from the low-level teacher policy to handle noisy observations following~\cite{miki2022wild}.

We then train the high-level teacher policy, whose primary task is to generate 6 DOF commands for the low-level policy while going over rough terrain and avoiding overhanging obstacles in a procedurally generated terrains. This policy also employs reinforcement learning and uses spherical scans as shown in Fig.~\ref{fig:overview} to capture localized geometric data. These scans provide an omnidirectional view of the environment and inform the high-level policy's decision-making process.

Finally, we train a high-level student policy, which is intended for the robot's final deployment in the field. Unlike the high-level teacher policy, the student policy is trained using noisy observations simulated to resemble data from the robot's on-board sensors. Specifically, it employs noisy voxel data as a three-dimensional representation of the environment.

\subsection{Low-level locomotion policy}

We extend the perceptive locomotion pipeline presented in~\cite{miki2022wild} by adding three more command inputs: desired roll $\xi$, pitch $\theta$, and body height $h$ to allow the robot to have more control over its motion.
The low-level teacher policy is trained using \ac{PPO} algorithm~\cite{schulman2017proximal}, whereby the robot interacts with the environment and receives rewards based on its performance in following the given 6D commands on the rough terrain. 

The command is randomly sampled at the beginning of each episode, ensuring that the robot is exposed to a variety of command scenarios during training.
We sample roll and pitch command from a normal distribution with mean zero and standard deviation 0.25 rad and the desired body height from the ground is randomly sampled from an interval ranging from 0.1 to 0.6 above ground level. 
To avoid spikes in ground height values near edges, we use the average of five sampled ground heights to represent the body's height above the ground.

The low-level teacher observation includes a 6D command, proprioceptive observation $o_p^L$, exteroceptive observation $o_e^L$, and privileged state $s_p^L$. The proprioceptive observation ($o_p^L$) consists of the body velocity, orientation, history of joint position error and velocity 
 that stack few frames between each control loop, action history, as well as the phase of each leg. The exteroceptive observation ($o_{e}^{L}$) is represented by height samples taken around each foot with five distinct radii~\cite{miki2022wild}.
The privileged state $s_p^L$ encompasses various factors such as contact states, contact forces, contact normals, friction coefficient, thigh and shank contact states, external forces and torques applied to the body, and swing phase duration.
We adopt the same action space as that described in previous research~\cite{miki2022wild} which consists of the phase difference $\varDelta \phi_l$ of the periodic motion generator and the residual joint position target $\varDelta q_i$ that is added on top of the periodic motion.
To represent our teacher policy $\pi_{\theta}$ we utilize a Multi-Layer Perceptron model.

We add two additional reward terms on top of the original rewards which consists of velocity tracking rewards and penalties such as joint torque, velocity, acceleration or slip rewards. Please refer to the previous research~\cite{miki2022wild} for the detailed reward definitions for locomotion.
The additional reward terms are employed for roll, pitch and body height tracking. 

The orientation tracking reward is:
\begin{equation}
r_{\text{orientation}} = \exp(-\alpha \times e_{rp}),
\end{equation}
where \( e_{rp} \) denotes the error in either roll or pitch and \( \alpha \) is a scaling factor that determines the sensitivity of the reward to orientation errors.
Similarly, the body height tracking reward used to encourage the robot to maintain a desired body height is described as:
\begin{equation}
r_{\text{base height}} = \exp(-\alpha \times e_{h}),
\end{equation} 
where the base height error, \( e_{h} \), represents the absolute distance between the desired and actual body heights above the ground. 

The student policy for the low-level policy is trained following the same manner of the previous work~\cite{miki2022wild} to remove the privileged observation dependency and apply noise on the exteroception.
User
It was trained using extensive domain randomization applied to exteroceptive measurements, employing behavior cloning to minimize the action difference between the teacher policy.

\subsection{High-level teacher policy}
After we get the low-level policy, a high-level teacher policy is trained with \ac{PPO} to command the low-level policy in navigating diverse terrains that include overhanging obstacles. While the low-level policy focuses on training specifically for rough terrain, the high-level policy is trained in confined spaces with overhanging obstacles. 
This separation in training allows us to focus the high-level policy training on the specific challenges of navigating confined spaces.

The high-level policy's task is to follow a desired velocity target by commanding the low-level policy.
It takes proprioception $o_p^H$ as well as exteroceptive observations $o_e^H$ as input. 
$o_p^H$ includes velocity commands, body velocities, joint positions, joint velocities, body orientation, and previous actions. $o_e^H$ consists with the same height samples as the low-level policy ($o_e^L$), and a spherical scans for omnidirectional observations.
We used the sparse ray tracing pattern, because using more complicated data structure, such as depth images or voxel grids, could slow down the reinforcement learning process due to their computational requirements.
As the action space, a residual approach was utilized to increase training efficiency where it outputs the residual for $v_x$, $v_y$ and $\omega_z$ instead of outputting the velocity commands directly. This can skip the process of learning to output the same input velocity commands to track them. For commands related to roll ($a_{\xi}$), pitch ($a_{\theta}$), and base height ($a_h$), they are directly outputted. 

We employ rewards consisting of task rewards that encourage the desired behavior and penalty rewards to avoid undesirable behavior. For task rewards, we used velocity tracking reward and base distance reward, designed to maintain a safe distance from obstacles and thus enhance safety during navigation.
For base distance reward we used 
\begin{equation}
    r_{base} = \exp(-\alpha \times (d_{max} - \min(d, d_{max}))),
\end{equation}
where $d_{max}$ is a parameter to set a threshold and $\alpha$ defines the coefficient. To measure the distances $d$, we used ray casting in a spherical pattern.
For penalties, we discourage the robot from hitting objects with a collision penalty by simply giving a large negative reward when the body hits obstacles. We also use penalties for joint speed, acceleration and torque for smoother motion and vertical velocity and orientation to avoid unnecessary movements.

\subsection{High-level student policy training}
After training a teacher policy that can perform the task using ground truth data, we distill this policy into a student policy that can be deployed within the sensory constraints of a physical robot.
We convert the spatial information into the occupancy voxel representation as an abstraction. This allows to combine different sensor configurations and also allows using voxel mapping pipeline~\cite{oleynikova2017voxblox,octomap,Hoeller2022-fw}, or transforming the past observations into the current frame~\cite{Yang2023-tm} to accumulate information.

To compress the voxel representation to a smaller latent space before feeding to the belief encoder, we leverage a three layer voxel encoder to compress the data into a more compact latent space. The voxel grid dimensions are 32×32×32, with a resolution of 0.08 along each axis. The voxel encoder consists of 3D convolutions coupled with layer normalization and ELU activation following the methodology described in~\cite{Hoeller2022-fw}. 

We use a belief encoder based on a Gated Recurrent Unit (GRU) to integrate the proprioceptive data and the external latent representation into a coherent belief state similar to~\cite{miki2022wild}. We flattened the voxel latent representation before putting into the belief encoder. The belief encoder has a hidden size of 128 and consists of two layers.
Additionally, we incorporate a belief decoder that decodes privileged information from the latent feature to help extracting information that is useful for student policy.
Finally, the output of the belief encoder is fed into an MLP network, consistent with existing architectures, to produce the final action commands.
The loss is the action loss which is the mean square error between the teacher policy and the reconstruction loss of the privileged states.

\subsection{Procedural Terrain Generation}

\begin{figure}[htbp]
\centering
\includegraphics[width=\linewidth]{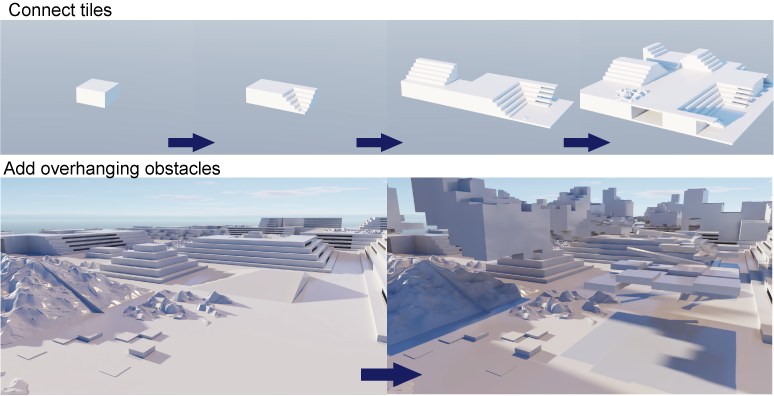}
\caption{Procedural terrain generation. We generate the terrain mesh by tiling mesh parts. First, the tiles are connected procedurally based on connectivity calculated from terrain height array. Then, overhanging obstacles are added on top of the mesh.}
\label{fig:terrain_generation}
\end{figure}
To simulate different environments, we developed a procedural terrain generator that builds mesh representations of different terrains\footnote{We open-sourced the software \url{https://github.com/leggedrobotics/terrain-generator}}. This generator uses the \ac{WFC} method~\cite{wfc}, widely used in game development, to ensure smooth transitions between neighboring tiles based on their connectivity. We used terrain heights to calculate the connectivity between each tile.

We defined \textit{1570} types of tiles, including terrains such as steps, stairs, ramps and rough in different heights. After the mesh was generated, boxes of different dimensions were added to the terrain to simulate confined spaces. This method allows for the generation of a wide variety of terrain combinations, reflecting the complex environments a robot may encounter, as shown in Figure \ref{fig:terrain_generation}.

\begin{figure*}[ht]
\centering
\includegraphics[width=0.9\linewidth]{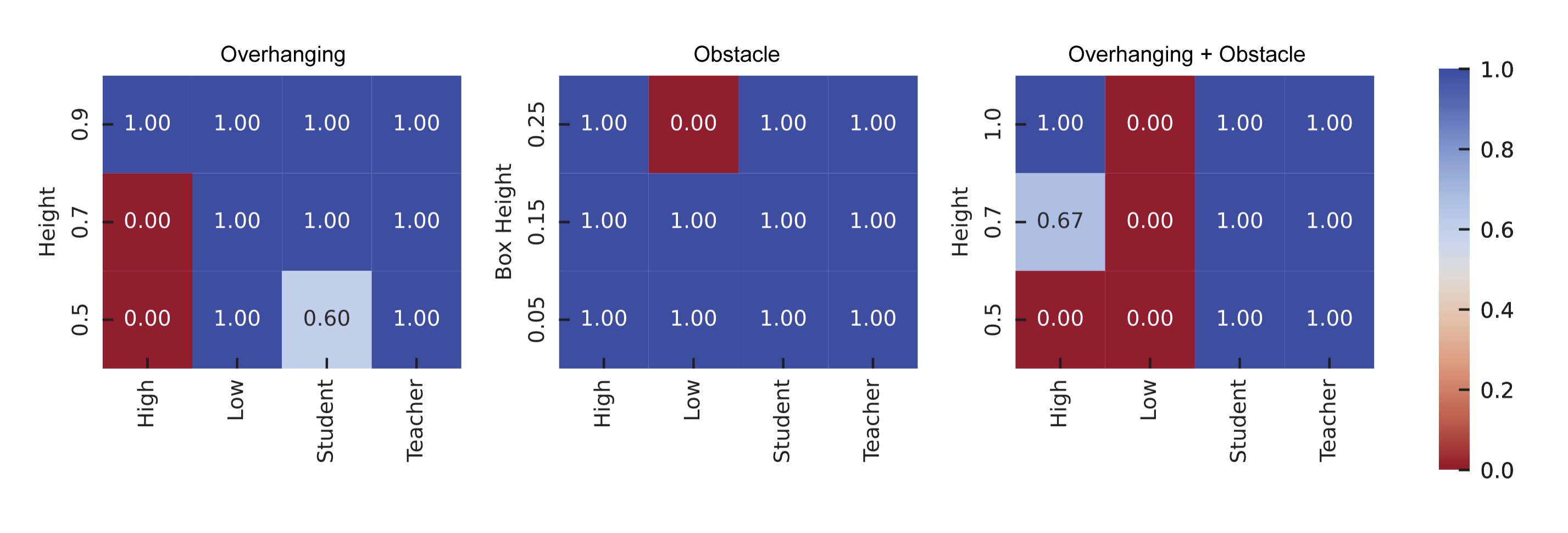}
\caption{Success rate of different overhanging obstacle height and obstacle box height from the ground. The $x$ axis show different methods and $y$ axis shows the different parameters of the obstacles. For the obstacle + overhanging, we used the obstacle height of $0.25$m and varied the height of the overhanging box. The baseline methods which always walk at normal height (High) and always walk with crouching (Low) was compared against our method. The results show that the combination of overhanging and rough terrain needs an adaptive body height control.
}
\label{fig:success rate heat}
\end{figure*}

\section{Experimental Results}
\begin{figure}[ht]
\centering
\includegraphics[width=0.75\linewidth]{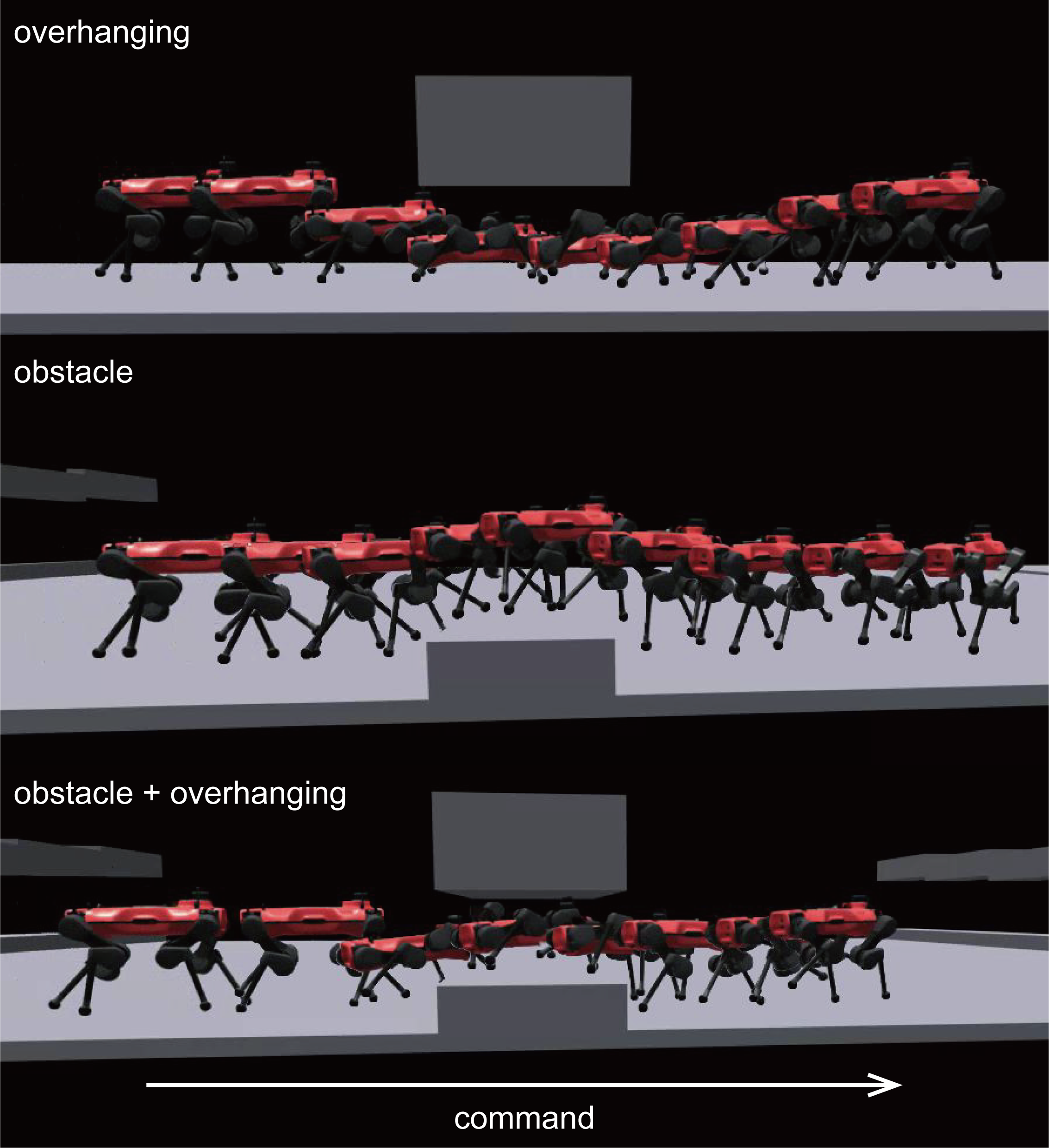}
\caption{Sequence on evaluation terrains. We have evaluated the policy's performance on the evaluation terrains where it has three terrain types with different parameters. Overhanging has an overhanging box in the middle, and Obstacle has a box on the ground, Overhanging + obstacle is a combination of both obstacle and overhanging boxes.}
\label{fig:terrain_types}
\end{figure}

\begin{figure}[htbp]
\centering
\includegraphics[width=\linewidth]{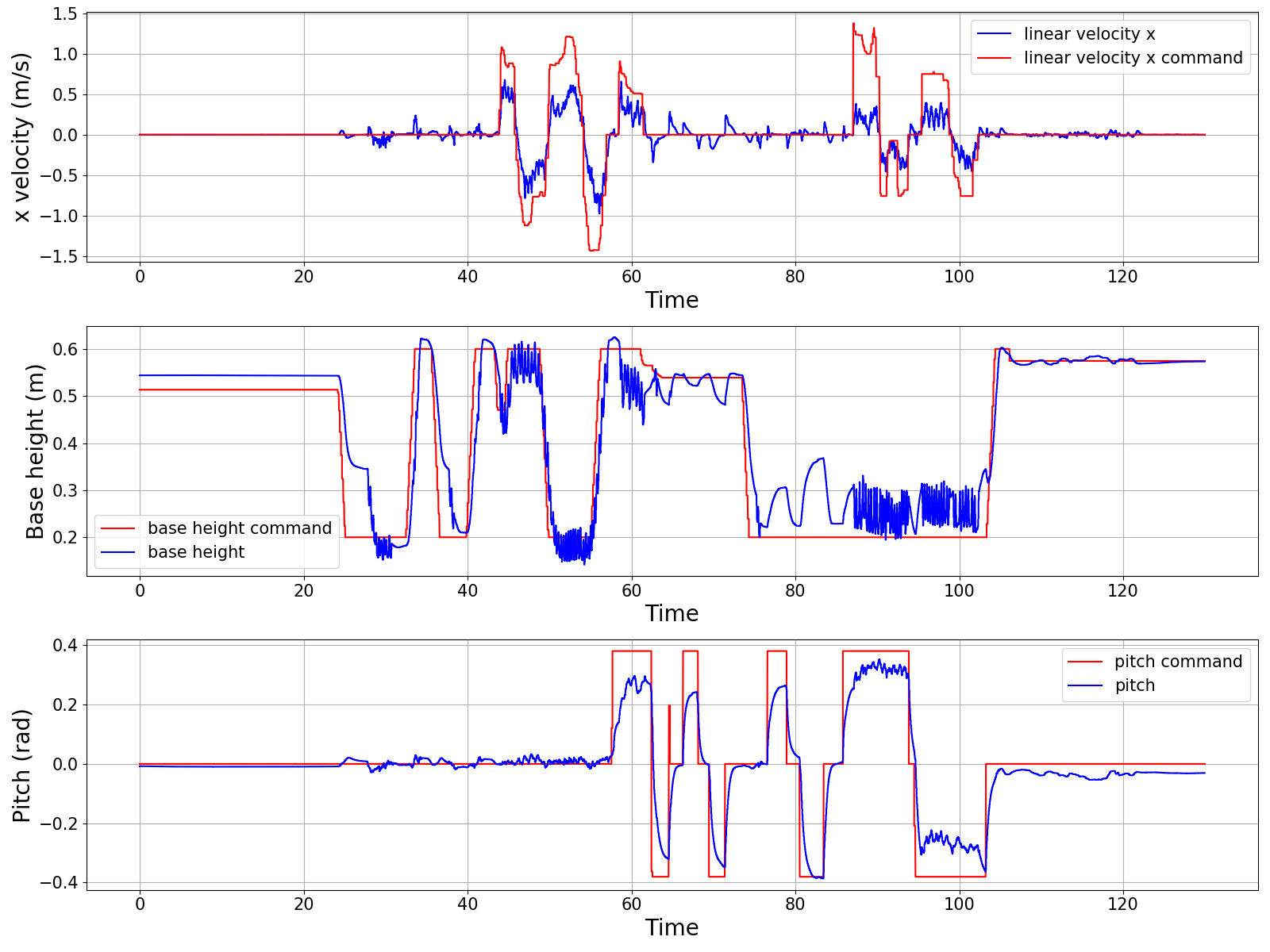}
\caption{Tracking performance of a low-level policy when responding to simultaneous commands for body height, pitch, and linear x velocity. The results confirm that the policy can effectively manage combined actions such as crouching, tilting, and walking simultaneously.}
\label{fig:low-level-policy-plot}
\end{figure}

\begin{figure*}[htbp]
\centering
\includegraphics[width=\linewidth]{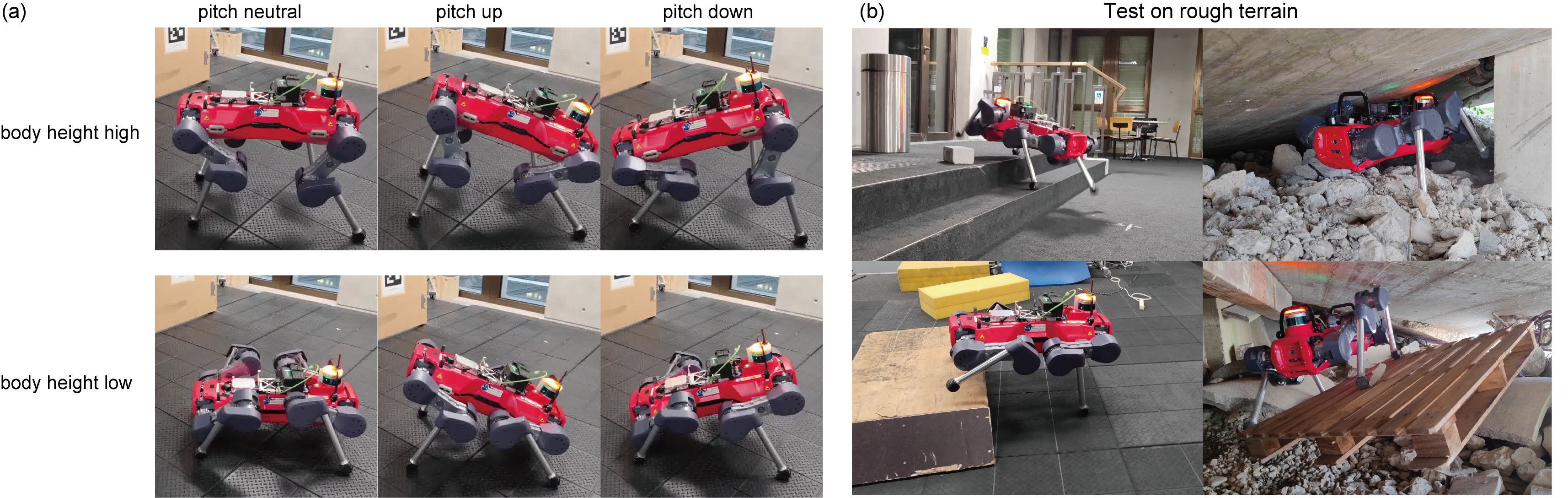}
\caption{Tests of the low-level policy on the hardware. (a) The robot could follow additional commands such as body height or pitch. (b) The policy showed its robustness on different rough terrains such as stairs, step, loose ground or wooden steps while following low body height command.}
\label{fig:low-level-policy}
\end{figure*}

\begin{figure*}[ht]
\centering
\includegraphics[width=\textwidth]{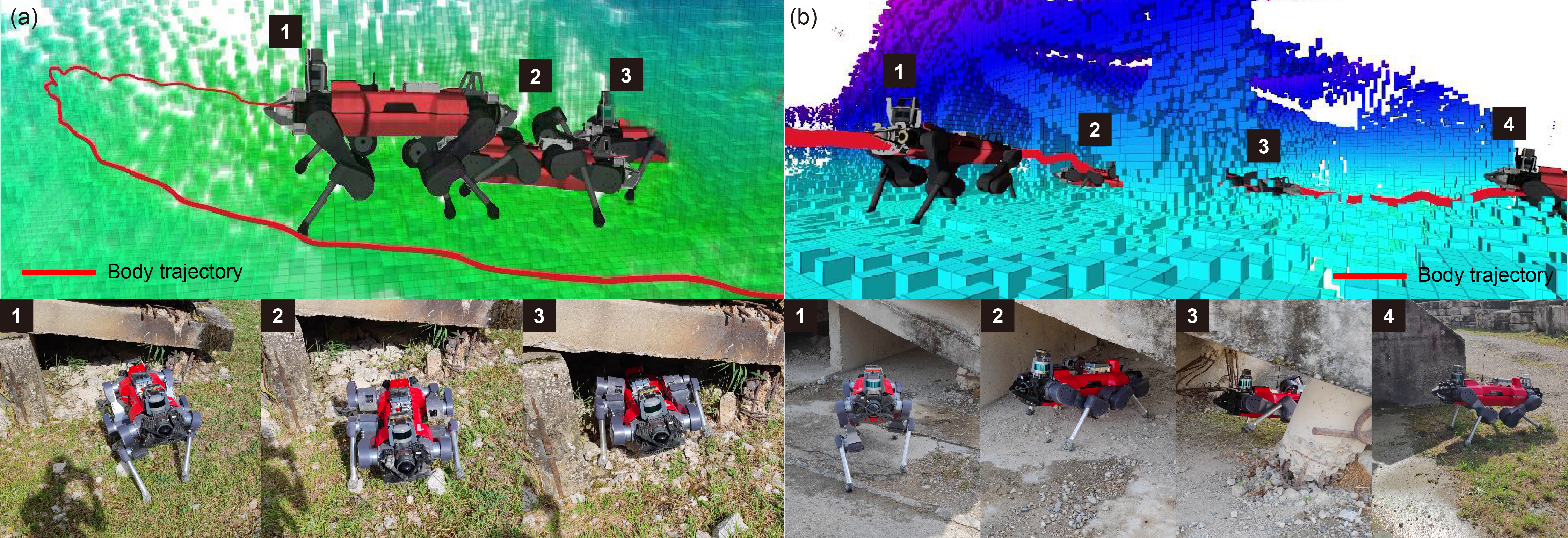}
\caption{Real-world deployment of the integrated high-level and low-level policies. The sequence of images illustrates the robot's ability to autonomously adapt its body height in the presence of overhanging obstacles.}
\label{fig:experiment_confined}
\end{figure*}

In this section, we describe the setup of our experiments, which include both simulations and real-world tests. In the simulation phase, we evaluate the performance of the policy on evaluation terrains. After that, we test the trained policy on actual hardware to validate the performance in real-world.

\subsection{Evalution in simulation}
To evaluate the high-level student policy, we created a test environment with a variety of obstacle heights and overhanging obstacles. In this setup, the robot had to move from a starting position to a goal position that is 6 meters away. The criterion for a successful trial was the robot's ability to reach the target within 30 seconds. Note that we applied these test conditions to student policies that had been trained on procedurally generated terrains, not to policies specifically trained on the evaluation setup.  The results are detailed in Figure \ref{fig:success rate heat}. 
We compared the performance against two baselines both employ simple strategy without the high-level policy. One is the policy which always walk at normal height (High), and the another one always crouching on the ground (Low). 
The result shows that the student policy could handle three types of the terrains fairly well and closely matched the performance of the teacher policy. On the other hand, the Low baseline could traverse overhanging obstacles but failed to overcome obstacle on the ground and High baseline could traverse obstacles on the ground but failed to go through overhanging obstacle.

\subsection{Real-World Experiments}
Our trained policies were empirically tested using ANYmal-C and ANYmal-D robots equipped with either dome LiDARs or depth cameras. 
Depth sensor data were transformed into voxel structures - achieved by voxel mapping for depth cameras and straightforward voxelization in the LiDAR configuration. 

\subsubsection{Low-level policy}
First, we evaluate the low-level policy's capabilities with the operator giving commands to guide the robot's behavior.
The robot effectively followed the combinations of different commands as shown in Figure \ref{fig:low-level-policy-plot}.
Even under complex conditions, such as maintaining a crouched position while traversing stairs, steps and uneven terrain, the robot was able to adjust its body height and orientation according to the operator's instructions. These empirical results, illustrated in Figure \ref{fig:low-level-policy}, show the robustness and adaptability of our low-level policy.

\subsubsection{Combined policy in the field}
In the final part of our experiments, we tested a combined high-level and low-level policy in a real-world environment, as shown in Figure \ref{fig:deployment} and \ref{fig:experiment_confined}. The test environment consisted of a simulated collapsed building and some objects such as tables and benches.

In the tests, the operator provided commands for the robot's $x$, $y$, and yaw velocities and the robot autonomously determined its own body height, roll, and pitch.
The robot faced several challenges, such as navigating confined spaces with ceilings at varying angles. The terrain conditions were also difficult, with loose and rough gravels and unstable steps. However the robot could handle all challenges robustly and showed that the robot could reliably enter and traverse confined spaces under challenging conditions. This confirms the robustness of our combined policy approach in real-world scenarios.

\subsection{Simulation details}
For low-level policy training, we used RaiSim~\cite{raisim} to simulate the robot and the environment, while 
for high-level policy, Isaac Gym~\cite{isaacgym-2021} was used to simulate confined space environment with depth sensor measurements such as spherical scan, depth camera and voxel map.
We used 1000 parallel environments for the low-level and high-level teacher policy training and 300 for the student policy training.

\section{CONCLUSIONS}
In this paper, we present a legged robot system designed for navigating confined spaces, broadening the operational environments for robotics. Our framework adopts a two-layer hierarchical policy structure to decompose the complexity of navigation into more manageable components. The low-level policy ensures robust traversal over uneven terrain based on 6D commands, including body velocity and orientation, while the high-level policy strategically directs these commands using 3D spatial understanding. Our validation involved training on diverse terrains and deploying in a challenging collapsed building scenario, demonstrating the system's real-world applicability. However, limitations exist in highly dynamic environments and performing advanced maneuvers like using body contacts in tight spaces, pointing to future research directions.

\bibliographystyle{IEEEtran}
\bibliography{reference.bib}

\end{document}